\pdfoutput=1

\documentclass[11pt]{article}

\usepackage[final]{acl}

\usepackage{times}
\usepackage{latexsym}
\usepackage{amsmath}
\usepackage{adjustbox}

\usepackage[T1]{fontenc}

\usepackage[utf8]{inputenc}

\usepackage{microtype}

\usepackage{inconsolata}

\usepackage{graphicx}

\usepackage{listings}
\usepackage{xcolor}
\usepackage{helvet}
\title{Large Language Models for Mathematical Analysis}

\author{Ziye Chen \\
  Boston University \\
  \href{mailto:ziyechen@bu.edu}{\texttt{ziyechen@bu.edu}} \\\And
  Hao Qi \\
  Boston University \\
  \href{mailto:haoqi@bu.edu}{\texttt{haoqi@bu.edu}}
  }

\begin{document}

\maketitle
\begin{abstract}

Mathematical problem-solving is a key field in artificial intelligence (AI) and a critical benchmark for evaluating the capabilities of large language models (LLMs). While extensive research has focused on mathematical problem-solving, most existing work and datasets concentrate on computational tasks, leaving gaps in areas like mathematical analysis, which demands rigorous proofs and formal reasoning. We developed the \texttt{DEMI-MathAnalysis} dataset, comprising proof-based problems from mathematical analysis topics such as Sequences and Limits, Infinite Series, and Convex Functions. We also designed a guiding framework to rigorously enhance LLMs' ability to solve these problems. Through fine-tuning LLMs on this dataset and employing our framework, we observed significant improvements in their capability to generate logical, complete, and elegant proofs. This work addresses critical gaps in mathematical reasoning and contributes to advancing trustworthy AI capable of handling formalized mathematical language. The code is publicly accessible at \href{https://github.com/ziye2chen/LLMs-for-Mathematical-Analysis}{LLMs for Mathematical Analysis}. 

\end{abstract}


\section{Introduction}

Mathematical analysis, with its emphasis on rigorous proofs and formal methods like the $\epsilon$-$\delta$ definition of limits, poses a significant challenge for artificial intelligence (AI). Large language models (LLMs) have shown remarkable advancements in solving computational problems across various domains, yet they often struggle with the formal rigor and reasoning required for mathematical analysis. Generating correct and structured solutions in this field remains a challenging task, as LLMs frequently rely on uncritical, computational shortcuts or fail to produce logically sound proofs. (See Appendix \ref{appendix:gptChallenge} for a test example.) 

Existing mathematical datasets for fine-tuning and benchmarking LLMs primarily focus on computational tasks in areas like algebra, calculus, and geometry, while deliberately avoiding proof-based problems. This limitation hinders the development of LLMs capable of solving problems that require precise reasoning and formal processes.

To address this gap, we developed the \texttt{DEMI-\\MathAnalysis} dataset, a specialized corpus of proof-based problems in real analysis sourced from \textit{Problems in Mathematical Analysis} \cite{demidovich1964problems} and \textit{Problems and Solutions in Real Analysis} \cite{hata2007problems}. This dataset includes diverse topics such as Sequences and Limits, Infinite Series, and Convex Functions. Additionally, we designed a guiding framework to improve LLMs' ability to generate rigorous, clear, and logically sound solutions.

By fine-tuning models like Llama 3.2 \cite{dubey2024llama3herdmodels} and Qwen2 \cite{yang2024qwen2technicalreport} on this dataset and applying the framework, our work advanced the field of AI-facilitated reasoning and contributed to building trustworthy AI capable of handling complex, formalized mathematical languages.

\section{Related Work}

\subsection{Mathematics Benchmarks for AI}

\begin{figure}[htbp]
  \includegraphics[width=\columnwidth]{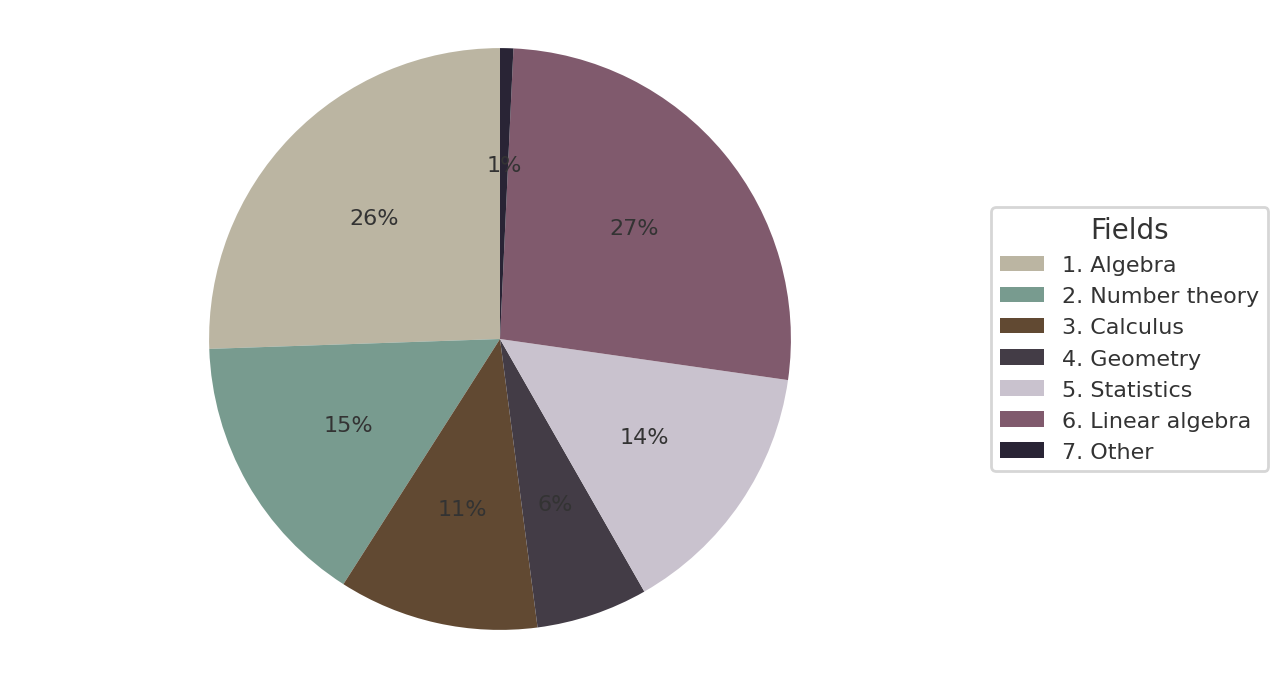}
  \caption{Mathematical fields distribution of current
datasets. Note that most of the questions are computation-related with a finite answer.}
  \label{fig:pie}
\end{figure}

The rapid development of AI has prompted the continuous introduction of mathematics benchmarks. GSM8K \cite{cobbe2021trainingverifierssolvemath} is a dataset of grade school math word questions, while MATH \cite{hendrycksmath2021} contains challenging competition problems. MathVerse \cite{zhang2024mathversedoesmultimodalllm} collected multi-subject problems with diagrams, and GeoEval \cite{zhang2024geoevalbenchmarkevaluatingllms} facilitated a deeper investigation into the performance of LLMs in solving geometry problems. TAL-SCQ5K \footnote{\url{https://github.com/math-eval/TAL-SCQ5K}} consists of multiple-choice questions in both English and Chinese. We collect and present the distribution of mainstream dataset topics in Figure \ref{fig:pie}.

\subsection{LLMs for Mathematics}

As extensive research progresses, LLMs have demonstrated proficiency in solving mathematical problems. \citealp{53097} introduced AlphaGeometry, a theorem prover for Euclidean plane geometry that approached the performance of an average International Mathematical Olympiad gold medallist using a neuro-symbolic system. Studies by \citealp{wang2023selfconsistencyimproveschainthought} show that LLMs with over 100B parameters are capable of addressing intricate tasks by employing a chain-of-thought (CoT) \cite{wei2023chainofthoughtpromptingelicitsreasoning} when given a limited set of reasoning examples. And CoT frameworks have been used to enhance mathematical performance, incorporating other tools \cite{heyueya2023solvingmathwordproblems}. Researchers in OpenAI claimed that their newest model \cite{OpenAIo1} placed among the top 500 students in the U.S. in a qualifier for the USA Math Olympiad. However, some of the excellent performance on these benchmarks may result from memorizing the data from these benchmarks \cite{xu2024benchmarkingbenchmarkleakagelarge}. 

In the studies above, the impressive performance of LLMs is often shown by solving problems from fields like geometry or algebra since they can quickly be evaluated by checking the result. However, the importance of the solution process is ignored. We focus on the analyzing process to unlock the more comprehensive language potential of state-of-art models.  

\section{Motivation}

LLMs rely on extensive datasets to develop their reasoning and problem-solving capabilities. However, most existing mathematical datasets for fine-tuning and benchmarking LLMs focus on computational tasks in domains such as algebra, number theory, calculus, geometry, statistics, and linear algebra. These datasets almost exclusively contain questions that require numerical or symbolic computation while deliberately avoiding proof-based problems. This leaves a significant gap in the ability of LLMs to handle rigorous reasoning and formal proofs, especially in mathematical analysis.

Mathematical analysis requires a deep understanding of formal processes, such as $\epsilon$-$\delta$ proofs and precise logical reasoning. Current LLMs often fail to produce solutions that adhere to these rigorous standards, as they are trained primarily on datasets that emphasize computation over formal derivation. To address this limitation, we developed the \texttt{DEMI-MathAnalysis} dataset, which focuses on proof-based problems. By incorporating problems that emphasize rigor and logical progression, this dataset aims to equip LLMs with the tools needed to handle formalized mathematical reasoning. This work is motivated by the goal of enhancing LLMs' performance in domains that demand rigorous use of languages, ultimately contributing to the development of more trustworthy and capable AI systems. Specifically, this study aims to reach the following research objectives:

\begin{itemize} 
    \item $RQ_1$: How can we develop a dataset for pretraining and benchmarking LLMs on mathematical analysis? 
    \item $RQ_2$: How can we create a framework to improve LLMs' ability to solve mathematical analysis problems? 
    \item $RQ_3$: How can we effectively evaluate the solutions generated by LLMs to ensure correctness and rigor? 
\end{itemize}

\section{Dataset}

Given that there has been no benchmark for evaluating mathematical proofs, the \texttt{DEMI-MathAnalysis} dataset is designed to enhance the ability of LLMs to solve mathematical analysis problems using rigorous and formally restricted methods, particularly the $\epsilon$-$\delta$ technique. The dataset is divided into two parts: one for pretraining and one for benchmarking. Files can be founded at \href{https://github.com/ziye2chen/DEMI-MathAnalysis}{DEMI-MathAnalysis}. 

\subsection{Dataset Creation} 

\begin{figure}[htbp]
  \includegraphics[width=\columnwidth]{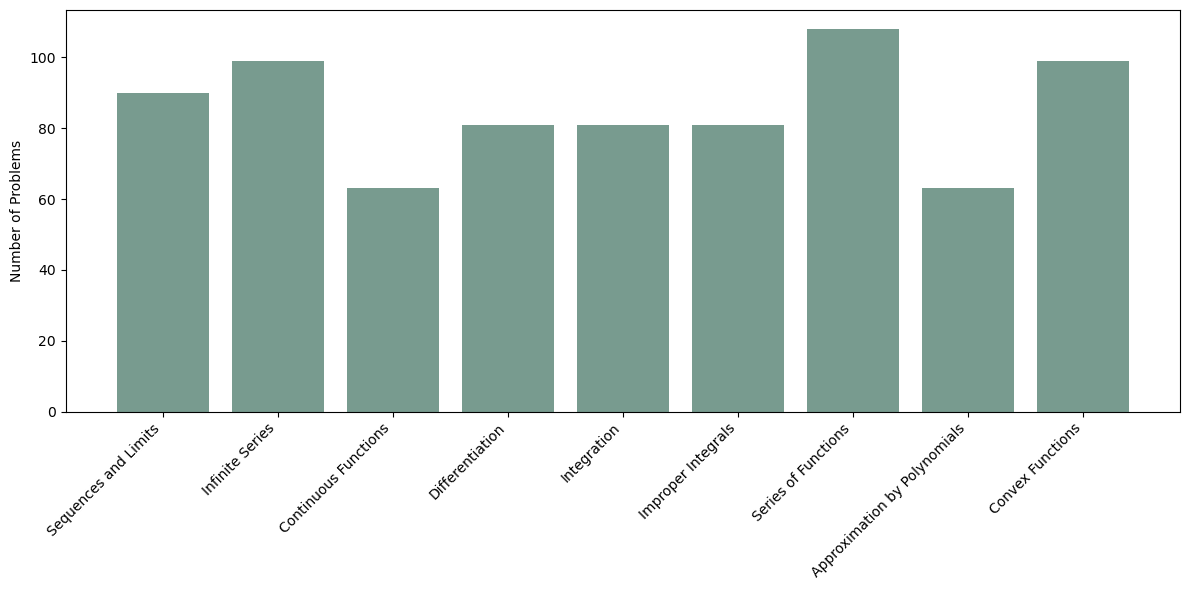}
  \caption{Number of questions per topic in \texttt{DEMI-
MathAnalysis}.}
  \label{fig:histogram}
\end{figure}

\texttt{DEMI-MathAnalysis} is built from problems in renowned texts. It covers topics like Sequences and Limits, Infinite Series, Continuous Functions, Differentiation, Integration, Improper Integrals, Series of Functions, Approximation by Polynomials, and Convex Functions. We ensure that topics of questions are distributed as evenly as possible and present the statistics in Figure \ref{fig:histogram}. 

\begin{figure}[htbp]
  \includegraphics[width=\columnwidth]{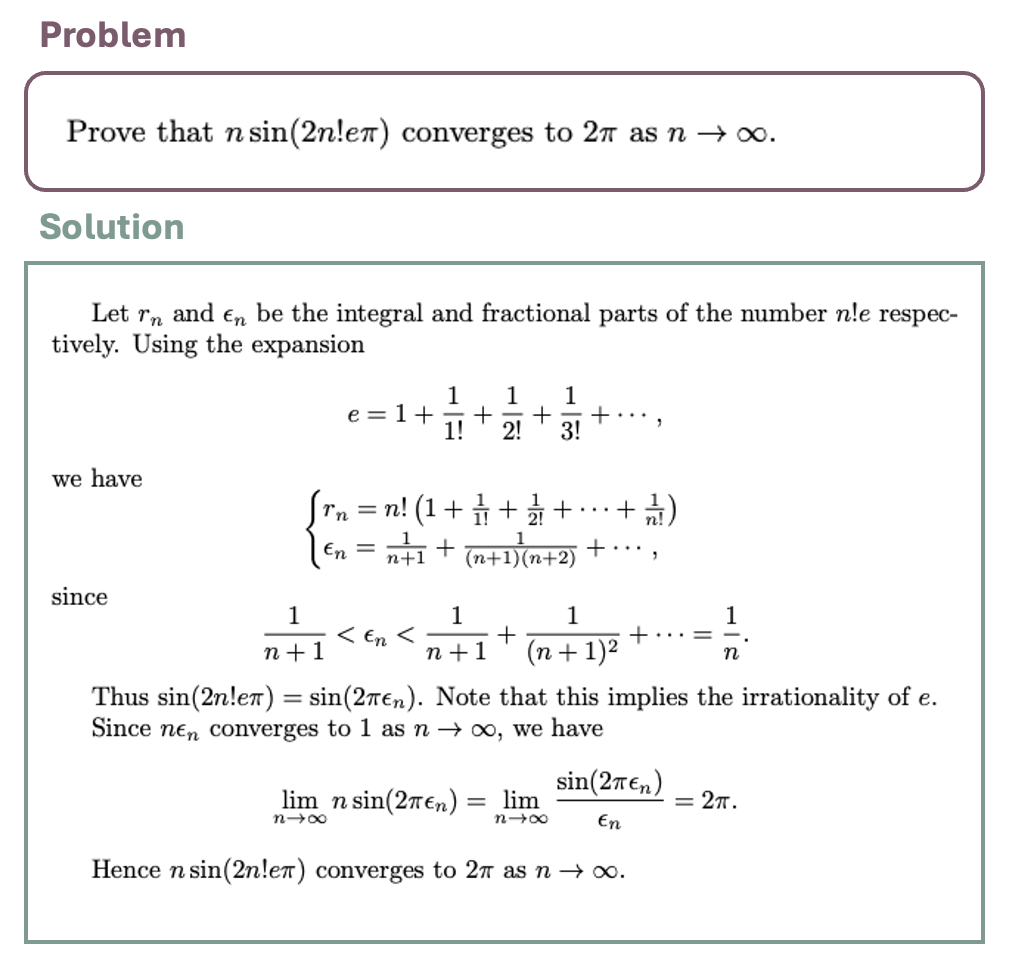}
  \caption{An example in \texttt{DEMI-MathAnalysis}. The LaTeX code has been rendered for better reading.}
  \label{fig:data}
\end{figure}

As an example in Figure \ref{fig:data} shows, each problem is transcribed in LaTeX format and paired with a detailed step-by-step solution. We have examined the quality of the answers to ensure that they do not include non-existent lemmas or incorrect grammar. 

\subsection{Dataset Structure} 

Each entry in the \texttt{DEMI-MathAnalysis} dataset consists of four components: \begin{enumerate} \item \textbf{Number:} A serial identifier linked to the original problem in the source material, enabling easy cross-referencing. \item \textbf{ProblemType:} A classification of the problem by its mathematical domain, aiding targeted training and evaluation. \item \textbf{Problem:} The problem statement, formatted in LaTeX to ensure clarity and precision for both humans and LLMs. \item \textbf{Solution:} A comprehensive, step-by-step solution written with formal mathematical rigor, highlighting key reasoning steps and adhering to standardized notation. \end{enumerate}

This dataset serves as a foundation for fine-tuning LMMs in the following section. It bridges gaps in existing mathematical datasets that focus primarily on computational tasks, ensuring that models are exposed to diverse problem types and guided toward generating logically sound, clear, and complete solutions. 

\section{Guiding Framework}

Based on the dataset, we proposed a framework for mathematical analysis that integrates functional components to guide LLMs in solving problems in a human-like way. By combining problem classification, knowledge retrieval, and solution generation, the framework introduces an adaptable pipeline for addressing reasoning and formatting complexities. 

\subsection{Components}

\begin{figure}[htbp]
  \includegraphics[width=\columnwidth]{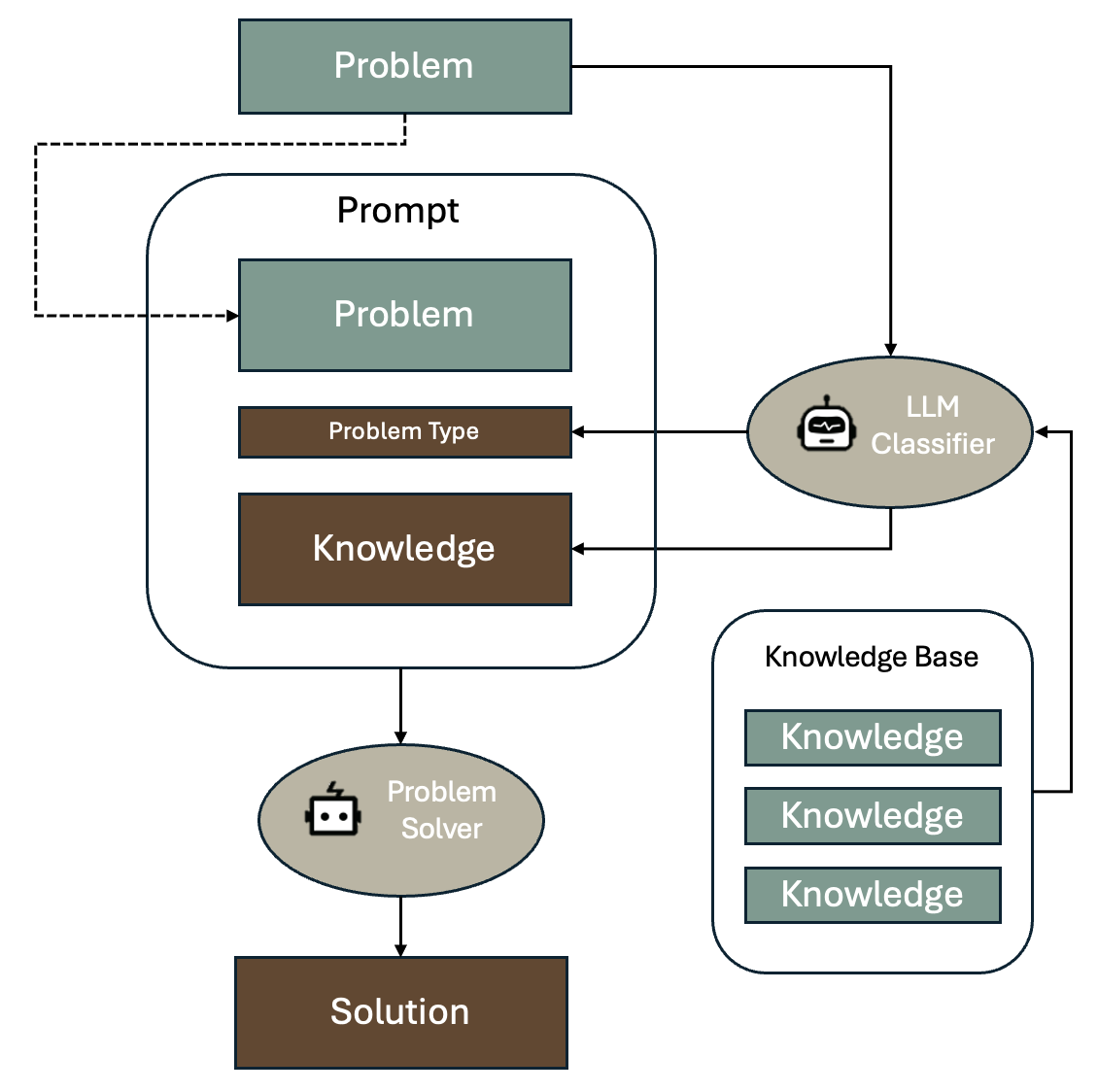}
  \caption{Framework of instructing analysis problems.}
  \label{fig:framework}
\end{figure}

As shown in Figure \ref{fig:framework}, the framework consists of the following key components:

\begin{enumerate} 

    \item \textbf{Problem Identification:} The input problem is first analyzed and classified into a specific category. This classification is performed by a lightweight LLM classifier trained on metadata from the \texttt{DEMI-MathAnalysis} dataset. Accurate classification ensures that the subsequent steps are tailored to the problem's mathematical domain, aligning the solution process with its requirements.

    \item \textbf{Prompt Construction:} Once the problem is classified, a detailed prompt is constructed to guide the LLM in generating a solution. The prompt includes:
    \begin{itemize}
        \item The full problem statement to provide complete context.
        \item The problem type, as determined by the classifier, to help the model focus on the appropriate reasoning approach.
        \item Supplementary knowledge retrieved dynamically from the Knowledge Base to provide relevant mathematical context, such as theorems, definitions, or key properties.
    \end{itemize}
    This process ensures that the LLM receives all necessary information in a structured format, optimizing its ability to process and solve the problem.

    \item \textbf{Knowledge Base Integration:} The Knowledge Base is a curated repository of mathematical concepts, rules, and formal methods specific to mathematical analysis. It includes:
    \begin{itemize}
        \item Key definitions, such as the $\epsilon$-$\delta$ definition of limits.
        \item Theorems and properties, such as those related to series convergence or convexity.
        \item Problem-specific heuristics, such as step-by-step methods for proving continuity or differentiability.
    \end{itemize}
    During prompt construction, relevant entries from the Knowledge Base are retrieved and incorporated into the prompt. This step ensures that the LLM is equipped with the required mathematical context, reducing reliance on general knowledge and increasing the rigor of the solution.

    \item \textbf{Solution Generation:}The Problem Solver module, powered by a fine-tuned LLM, uses the constructed prompt to generate a detailed solution. The solution generation process emphasizes:
    \begin{itemize}
        \item Logical rigor: Ensuring each step logically follows from the previous one.
        \item Completeness: Addressing all aspects of the problem and avoiding gaps in reasoning.
        \item Clarity: Presenting the solution in a well-organized and comprehensible manner.
    \end{itemize}
    The Problem Solver incorporates formal reasoning techniques, such as $\epsilon$-$\delta$ proofs, series approximations, and convexity arguments, to produce solutions that adhere to the rigorous standards of mathematical analysis.

\end{enumerate}

\subsection{Features and Benefits}

This framework bridges the gap between computational problem-solving and rigorous proof-based reasoning by equipping LLMs with the tools and methodologies needed to tackle formalized mathematical problems. It introduces several innovations to enhance LLMs' problem-solving ability, which requires powerful reasoning abilities:

\begin{itemize} \item Dynamic Prompt Adaptation: Prompts are dynamically tailored based on the problem type and retrieved knowledge, ensuring relevance and context-specific guidance. \item Fine-Tuned Models: The framework fine-tunes mainstream models on the \texttt{DEMI-MathAnalysis} dataset, optimizing them for proof-based thinking. \item Formal Reasoning Integration: The framework explicitly incorporates formal methods such as $\epsilon$-$\delta$ proofs and theorems on series convergence into the solution process.  \end{itemize}

\section{Experiment and Results}

To evaluate the effectiveness of our framework and provide a baseline of the \texttt{DEMI-MathAnalysis} dataset, we conducted extensive experiments using multiple language models. Our goal was to measure the improvements in solving proof-based problems in mathematical analysis, focusing on logical rigor, completeness, and clarity. 

\subsection{Experiment Setup}

We applied our framework to two relatively small models (Llama-3.2-3B and Qwen-2.5) and tested the state-of-the-art OpenAI o1-preview model. The prompts for fine-tuning and inference are listed in Appendix \ref{appendix:infer-prompt}. With the help of Unsloth \cite{han2023unsloth}, we fine-tuned the models faster with less memory cost. The hyper-parameter settings can be found in Appendix \ref{appendix:settings}. 

\subsection{Evaluation Setup}

\begin{figure}[htbp]
  \includegraphics[width=0.9\columnwidth]{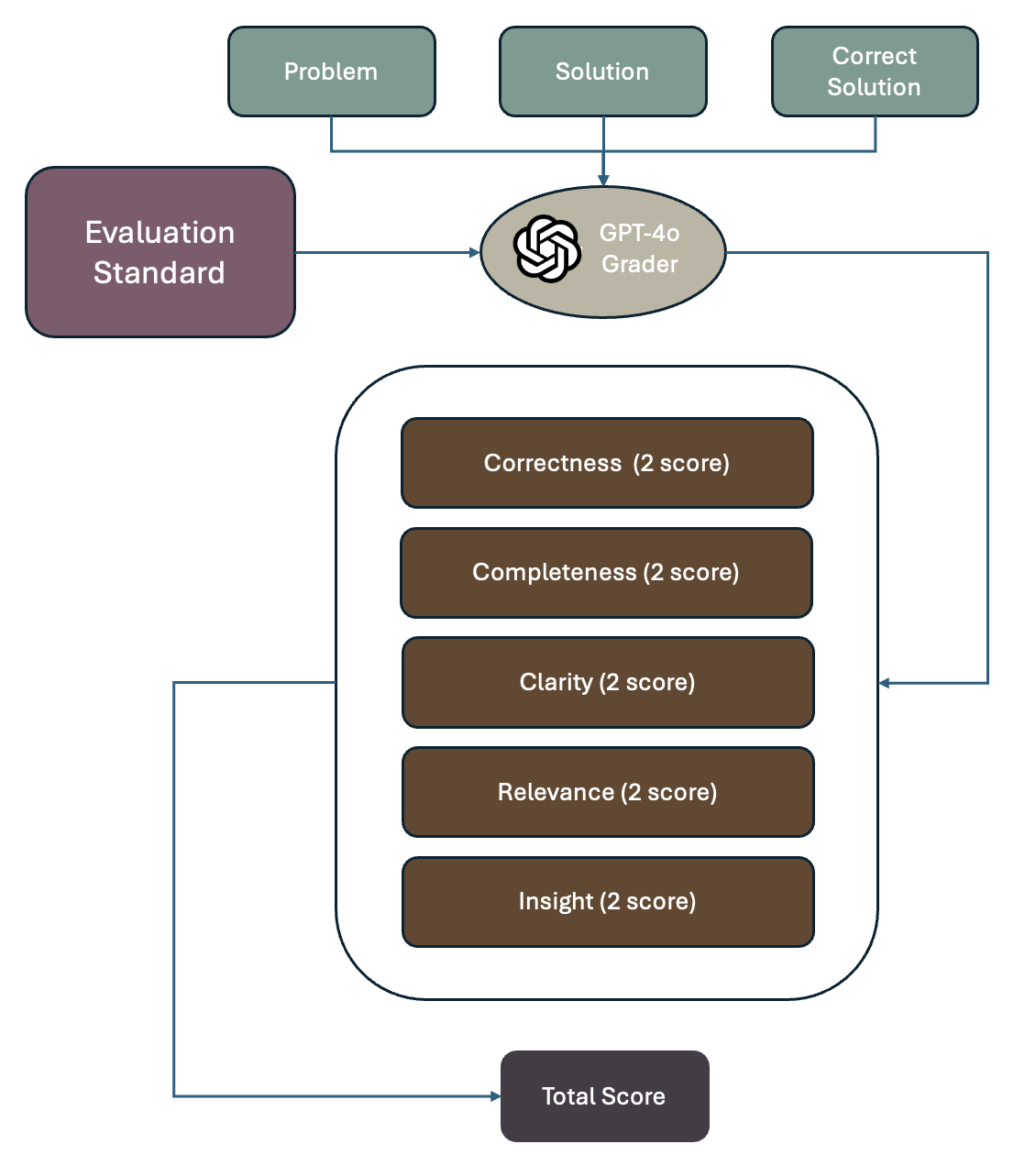}
  \caption{Proof evaluation process using GPT-4o.}
  \label{fig:evaluation}
\end{figure}

For evaluation, we utilized GPT-4o \cite{GPT4o} as an expert. The evaluation was based on five key indicators, with a total score of 10 points: 

\begin{itemize} 
    \item \textbf{Correctness:} Logical rigor and adherence to problem requirements. 
    \item \textbf{Completeness:} Full justification of all steps and handling of assumptions. 
    \item \textbf{Clarity:} Structured presentation and consistency in mathematical notation. 
    \item \textbf{Relevance:} Use of appropriate methods and avoidance of irrelevant details. 
    \item \textbf{Insight:} Understanding of concepts and elegance of the solution.
\end{itemize}

Figure \ref{fig:evaluation} shows the procedure, and the detailed prompt can be found in Appendix \ref{appendix:eval-prompt}. 

\subsection{Results and Discussion}

The results demonstrate significant improvements in the models’ ability to handle formal mathematical reasoning when fine-tuned with the dataset and the framework. Table \ref{tab:performance} summarizes the averaged evaluation scores for each model:

\begin{table}[htbp]
    \centering
    \begin{adjustbox}{width=\columnwidth}
        \begin{tabular}{lc}
            \hline
            \multicolumn{1}{c}{Model} & Averaged Score \\
            \hline
            Llama-3.2-3B-Instruct & 0 \\
            Fine-Tuned Llama-3.2 & 33.5\% \\
            Fine-Tuned Llama-3.2 with framework & \textbf{40.8\%} \\
            Qwen-2.5-Math-7B-bnb-4bit  & 0 \\
            Fine-Tuned Qwen-2.5 & 37.6\% \\
            Fine-Tuned Qwen-2.5 with framework & 38.6\% \\
            OpenAI o1-preview & 41.5\% \\
            \hline
        \end{tabular}
    \end{adjustbox}
    \caption{LLMs' performance on DEMI-MathAnalysis.}
    \label{tab:performance}
\end{table}

The evaluation revealed significant differences between baseline models, fine-tuned models, and models utilizing the framework. Both baseline models, Llama-3.2-3B-Instruct and Qwen-2.5-Math-7B-bnb-4bit, failed to handle the rigorous proof-based problems in the \texttt{DEMI-MathAnalysis} dataset, achieving an average score of 0. This result emphasizes the complexity of the dataset and the need for specialized fine-tuning. Fine-tuning alone brought substantial improvements, with Llama-3.2 achieving an average score of 33.5\% and Qwen-2.5 reaching 37.6\%.

Incorporating the proposed framework further enhanced performance. For Llama-3.2, the framework increased the score to 40.8\%, demonstrating its capability to guide models in generating more rigorous and logically sound solutions. Similarly, Qwen-2.5's performance improved to 38.6\%, showcasing the framework's adaptability across different models.

Comparatively, the OpenAI o1-preview model achieved the highest score of 41.5\%, underscoring its state-of-the-art capabilities. However, the results show that fine-tuning smaller models with the \texttt{DEMI-MathAnalysis} dataset and applying the framework allows these models to approach the performance of much larger systems.

These findings validate the effectiveness of our idea in enhancing LLMs' ability to tackle formal mathematical reasoning. The significant improvements observed after fine-tuning and framework integration demonstrate that even smaller models have the potential to achieve robust performance on proof-based problems when guided by structured methodologies.

\section{Conclusion}

This work addresses the challenges of solving mathematical analysis problems using LLMs by introducing the \texttt{DEMI-MathAnalysis} dataset and a novel framework designed to enhance their reasoning capabilities. The dataset fills a critical gap by focusing on rigorous, proof-based problems that are often absent in existing mathematical datasets. Complementing the dataset, the framework integrates problem classification, knowledge retrieval, and solution generation to guide LLMs toward producing logical, complete, and rigorous solutions. 

Evaluation results demonstrate significant performance improvements, particularly for fine-tuned models leveraging the framework. These results validate the effectiveness of the dataset and the framework in advancing LLMs’ ability to tackle proof-based reasoning tasks, bridging the gap between computational problem-solving and formal mathematical reasoning. However, one shortcoming comes that these results dominated by LLMs may fluctuate within a certain range. Without considering labor-intensive activities, a future task is to develop a more robust proof evaluation system. One can convert the outputs into Lean \cite{10.1007/978-3-030-79876-5_37}, a language that automates proofs, or design more detailed prompts.

Future work also involves expanding the dataset to include a broader range of mathematical topics and refining the framework for improved generalization and adaptability. By addressing these areas, we aim further to contribute to developing trustworthy and versatile AI systems. 

\bibliography{custom}

\appendix

\section{Current answer to a specific Problem from GPT-4o}
\label{appendix:gptChallenge}

If we input the following problem: 

\begin{quote}
    Prove that if $f(x)$ is a continuous function, then the function $F(x)=|f(x)|$ is also continuous. 
\end{quote} The answer for reference should be:

\begin{quote}
    Let $x_0$ be any continuous point, then for any $\epsilon > 0$, there always exists $\delta > 0$, such that when $|x-x_0|<\delta$, there is always $|f(x) - f(x_0)| < \epsilon$. From $$||f(x)| - |f(x_0)|| \le |f(x) - f(x_0)|$$ we know that$$||f(x)| - |f(x_0)|| < \epsilon$$ so $F(x)$ is continuous on $x_0$. And we can deduce that $F(x)$ is continuous since $x_0$ can be any continuous point.
\end{quote}

\begin{figure}[htbp]
  \includegraphics[width=\columnwidth]{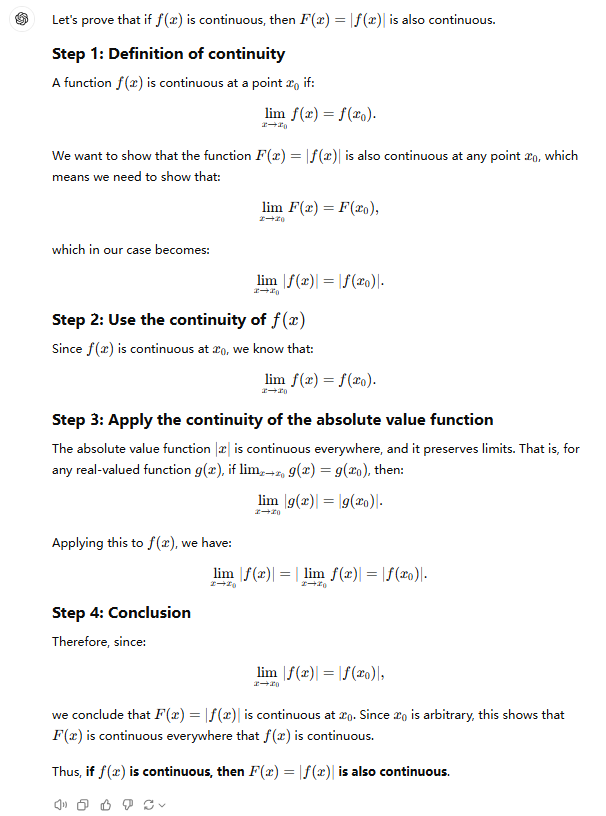}
  \caption{\textbf{Answer from GPT-4o without fine-tuning or guidance.} We can tell that it understands the connection between function limits and continuity, but it fails to provide a proof using the precise definition.}
  \label{fig:gptAnswer}
\end{figure}

Without fine-tuning and a guiding framework, the answer of LLMs, like GPT-4o, is shown in Figure \ref{fig:gptAnswer}, which is more cluttered and doesn't fit the rigorous mathematical language students typically learn about in a university course.

\section{Training settings}
\label{appendix:settings}

\begin{verbatim}
per_device_train_batch_size = 2,
gradient_accumulation_steps = 4,
warmup_steps = 5,
max_steps = 300,
learning_rate = 2e-4,
optim = "adamw_8bit",
weight_decay = 0.01,
lr_scheduler_type = "linear",
seed = 3407.
\end{verbatim}

\section{Prompt used for fine-tuning and inference}
\label{appendix:infer-prompt}

\subsection{Classifier}

\begin{lstlisting}
As a mathematical assistant, You need to analyze the problem to find out what type of problem it belongs to in Real Analysis. Provide the Problem_Type and the Knowledges which may be used to solve this problem.

### Problem:
{}

### Problem_Type:
{}

### Knowledge:
{}
\end{lstlisting}

\subsection{Solver}

\begin{lstlisting}
As a mathematical assistant, solve the following problem. Provide a detailed, step-by-step solution using rigorous mathematical reasoning. If the problem requires the use of the $\epsilon$-$\delta$ method (e.g., when proving limits or continuity), ensure that you apply it appropriately. Use precise mathematical language and notation throughout your solution.

### Problem_Type:
{}

### Problem:
{}

### Knowledge:
{}

### Solution:
{}
\end{lstlisting}

\section{Prompt used for evaluation}
\label{appendix:eval-prompt}

\begin{lstlisting}
You are a mathematical expert in Real Analysis. You need to evaluate the process of a proof of a real analysis problem. Please follow the steps to give a score to the solution:
Here are 5 key indicators to measure when evaluating a proof solution in Real Analysis:
1. Correctness
Logical Rigor: Does the solution logically follow from the given premises and known mathematical principles?
Adherence to Problem Requirements: Does the solution directly address the question and utilize the given conditions appropriately?
Accuracy: Are all mathematical statements, derivations, and conclusions valid?

2. Completeness
Full Proof: Is every step justified, leaving no significant gaps in reasoning?
Handling of Assumptions: Does the solution explicitly consider all assumptions and conditions stated in the problem?

3. Clarity
Structured Presentation: Is the solution organized logically, with clear progression from problem statement to conclusion?
Explanations: Are key steps, methods, and transitions explained clearly and concisely?
Notation Consistency: Is the mathematical notation consistent and aligned with standard practices?

4. Relevance
Appropriateness of Methods: Does the solution use the most relevant and efficient mathematical tools for the problem?
Avoidance of Irrelevant Details: Does the solution focus on solving the problem without introducing unnecessary complications or tangents?

5. Insight
Understanding of Concepts: Does the solution reflect a deep understanding of the underlying mathematical principles and the problem's nuances?
Elegance: If possible, is the solution concise and elegant, avoiding overcomplicated arguments?

Two points for each indicator for a total of ten points. And then I will give you the solution and the correct solution for your reference.
\end{lstlisting}

\end{document}